\documentclass[conference]{IEEEtran}
\IEEEoverridecommandlockouts
\usepackage{cite}
\usepackage{multirow}
\usepackage{amsmath,amssymb,amsfonts}
\usepackage{algorithmic}
\usepackage{graphicx}
\usepackage{textcomp}
\usepackage{xcolor}
\usepackage{listings}
\usepackage{booktabs}
\usepackage{url}
\usepackage{tabularx}

\def\BibTeX{{\rm B\kern-.05em{\sc i\kern-.025em b}\kern-.08em
    T\kern-.1667em\lower.7ex\hbox{E}\kern-.125emX}}
\begin{document}

\title{A Simulated Federated Analysis of MS-Induced Brain Lesions
}

\author{\IEEEauthorblockN{1\textsuperscript{st} Evelyn Trautmann}
\IEEEauthorblockA{\textit{Apheris AI GmbH} \\
Berlin, Germany \\
e.trautmann@apheris.com}
\and
\IEEEauthorblockN{2\textsuperscript{nd} Jo\"el Federer-Gsponer}
\IEEEauthorblockA{\textit{F. Hoffmann-La Roche Ltd.} \\
Basel, Switzerland \\
joel.federer-gsponer@roche.com}
\and
\IEEEauthorblockN{3\textsuperscript{rd} Markus C. Elze}
\IEEEauthorblockA{\textit{F. Hoffmann-La Roche Ltd.} \\
Basel, Switzerland \\
markus.elze@roche.com}
\and
\IEEEauthorblockN{4\textsuperscript{th} José-Tomás Prieto}
\IEEEauthorblockA{\textit{Apheris AI GmbH} \\
Berlin, Germany \\
j.prieto@apheris.com}
}
\maketitle

\begin{abstract}
    Federated techniques such as federated learning and federated analysis have emerged as a powerful paradigm for enabling multi-center research on sensitive clinical data while preserving patient privacy. In this study, we introduce a simulation framework that emulates a real-world federated research project focused on the analysis of multiple sclerosis (MS) patient data. The project comprises two components: an image segmentation task and a clinical data analysis task, where federated variants of survival analysis and Principal Component Analysis (PCA) are employed. To capture the complexity and heterogeneity of real clinical datasets, we construct a federation of high-fidelity synthetic cohorts designed to mirror MS-related clinical and demographic characteristics, while the imaging component leverages publicly available real-world datasets.

    Our simulation replicates key elements of authentic federated workflows, including distributed data governance, site-specific preprocessing, model training across isolated nodes, and the secure aggregation of analytical outputs. This framework provides a realistic testbed for developing, evaluating, and benchmarking federated learning methods in the context of MS research.
    
\end{abstract}
\section{Introduction}
This manuscript presents a simulated analysis of Multiple Sclerosis disease progression using both image and clinical data.
The analysis presented here is inspired by outcomes from the INTONATE network \cite{oh2023intonate, oh2023utility}.
The INTONATE-MS consortium is a public–private research consortium between Universitätsklinikum Münster, Penn Medicine, Unity Health Toronto, Erasmus MC and Roche. It constitutes a collaborative federated research framework that integrates large-scale, multicenter clinical trial data with real-world evidence (RWE) to enhance the understanding and management of multiple sclerosis (MS).
In particular, the image analysis follows the federated image segmentation study \cite{10.3389/fneur.2025.1620469} and the statistical part is based on the multi-center integration study \cite{oh2025rwPIRMA} as part of INTONATE.

In this paper, we demonstrate the interplay of image analysis, statistical inference, and survival analysis in a federated setting. We present an end-to-end workflow for multimodal, multi-center analysis that can provide a valuable contribution to drug development.

The image data used in this study comes from public datasets \cite{Guarnera2025}, while all clinical datasets are fully synthetic. Although the public image dataset includes clinical tables as well, we chose to generate artificial data to allow a wider variety of statistical methods and effects to be demonstrated.

The clinical data can be mapped to an OMOP CDM. Clinical measures, relapses and symptoms  can be aligned across observation and measurement tables, while demographics and disease attributes can be mapped from person, condition, and drug exposure tables.
A detailed mapping of source variables to their corresponding OMOP CDM tables is provided in Table~\ref{tab:omop-mapping}.

\begin{table}[ht]
\centering
\caption{Mapping of source variables to OMOP CDM tables.}
\label{tab:omop-mapping}
\begin{tabularx}{\columnwidth}{Xl}
\toprule
\textbf{Variable} & \textbf{OMOP CDM Table} \\
\midrule
PID, SEX, ETHNIC, BAGE & PERSON \\
VISITDT & VISIT\_OCCURRENCE \\
DIAGDT, MSSUBTP & CONDITION\_OCCURRENCE \\
TRTSDTC & DRUG\_EXPOSURE \\
DTFSTSYM & OBSERVATION\_PERIOD \\
RELAPSE, CDA, CNSR, & \multirow{3}{*}{OBSERVATION} \\
\quad VOCSTAT,EDUSTAT, & \\
\quad FOLLUPTM, PRSNTSYM & \\
EDSS, 9HPT, T25FWT, SDMT, & \multirow{3}{*}{MEASUREMENT} \\
\quad MSFC, LESION\_VOLUME, & \\
\quad  BASE, CHG  & \\
\bottomrule
\end{tabularx}
\end{table}
Some statistical patterns were intentionally designed for demonstration and do not reflect real clinical scenarios.
This paper demonstrates how federated statistics and federated machine learning can extract insights from data that are not directly accessible and distributed across multiple sites.
In particular, we demonstrate that a federated analysis on an entire statistical ensemble can reveal patterns that are not visible when considering only isolated subsets.

\section{Federated Analytics}
Federated analytics \cite{elkordy2023federated} is a computer-based system paradigm designed to enable joint analysis across distributed and sensitive medical datasets without requiring data centralization. In clinical and biomedical research settings, datasets are often siloed across institutions due to privacy, regulatory, and governance constraints, limiting the applicability of traditional centralized analytics pipelines. Federated analytics addresses this challenge by executing analytical computations locally at each data-holding site and sharing only intermediate results or aggregates for global analysis. 
Even though this approach requires a longer compute time, since the federated architecture entails additional communication overhead, the advantages are nevertheless evident, as data sources can be unlocked that would otherwise not be eligible for analysis.
From a computer-based medical systems perspective, the primary contribution lies in system architecture, orchestration, and secure computation. The approach does not focus on automated clinical decision-making. This makes federated analytics particularly suitable for multi-center studies, drug development, and medical research infrastructures.

\subsection{Architecture}
\begin{figure}
    \centering
    \includegraphics[width=\linewidth]{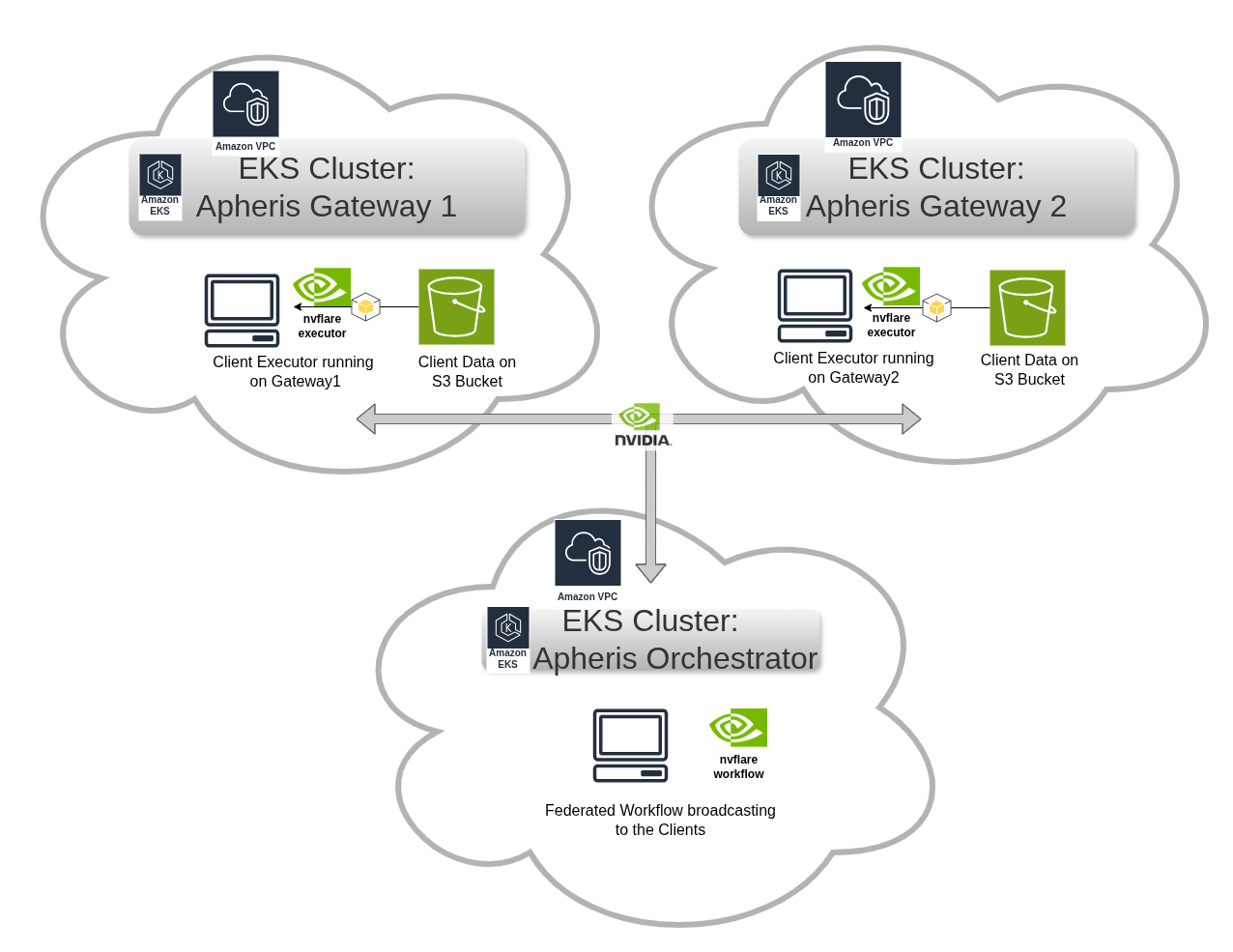}
    \caption{Apheris federated architecture \cite{AWS2025FederatedProteinLM}}
    \label{fig:fed-architecture}
\end{figure}

Figure \ref{fig:fed-architecture} illustrates the general architecture of a federated learning network as implemented by Apheris.

The Gateway is an agent that network participants can deploy into a Kubernetes cluster, which launches computations as pods in this cluster. Each Gateway, hosting its local data, deploys within its own isolated Virtual Private Cloud (VPC). The central Apheris Orchestrator, responsible for model parameter collection and aggregation, is also deployed in its own VPC.  

\begin{figure}
    \centering
    \includegraphics[width=\linewidth]{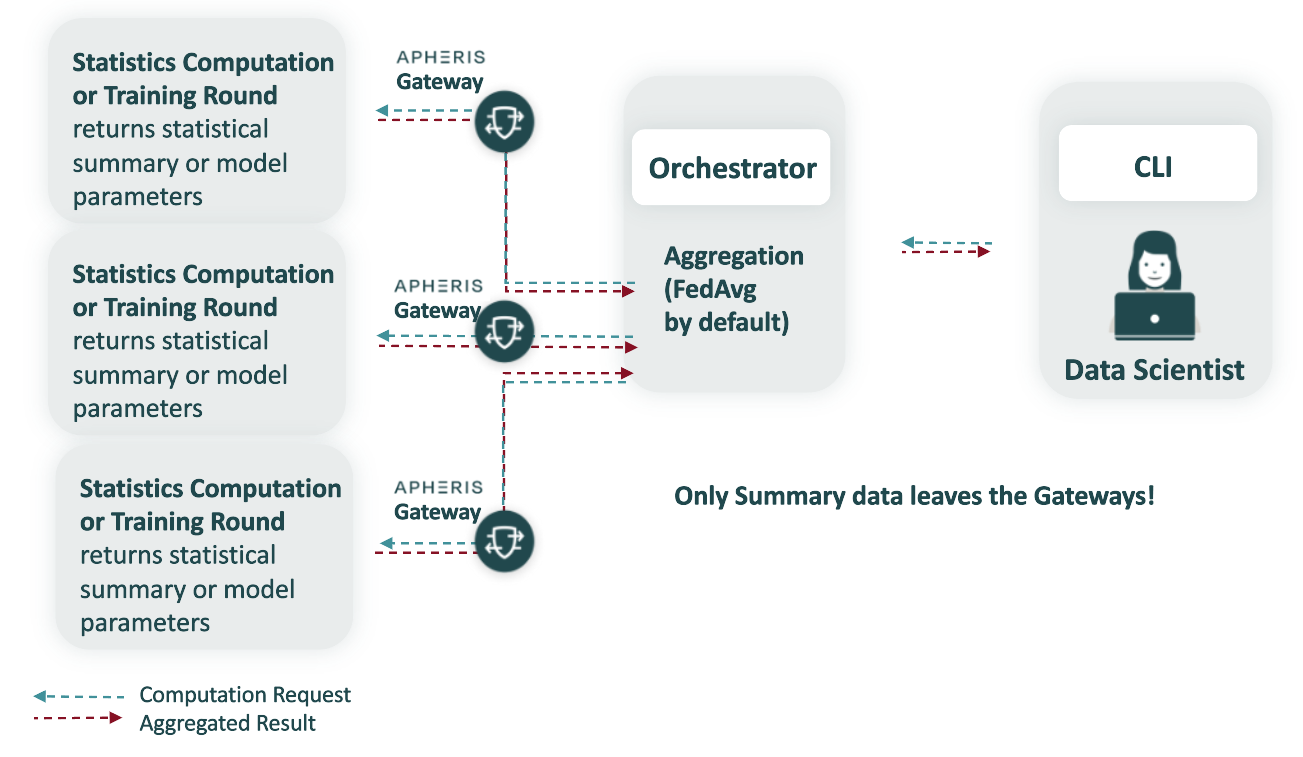}
    \caption{Federated computation workflow with central aggregation on the orchestrator.}
    \label{fig:fedcomp}
\end{figure}

Access to datasets registered to an Apheris Gateway and privacy controls are controlled by asset policies, ensuring that sensitive patient data can remain local while still contributing to global model development \cite{Apheris2025FederatedDataNetworks,HAGESTEDT2024101077}. 
An overview of the federated computation and training workflow is shown in Figure~\ref{fig:fedcomp}.

\subsection{Gateway-side Dataset Setup}

The analysis was designed around two synthetic sites, each containing both image and tabular data. The datasets were registered in each location to an Apheris Gateway, the procedure is detailed in the Apheris documentation \cite{Apheris_gateway}. The data remains secure, with asset policies granting access only to authorized compute jobs. Data remains at its original location; it is never downloaded or transmitted.
Each compute gateway is tied to a single organization. 
We use a setup with two gateways for the analysis, and to each of them a tabular clinical dataset and an image dataset are registered. The registration of a dataset to a compute gateway connects the data with the Apheris product. 

\subsection{Open Source Federated Frameworks}
Apheris' federated engine is based on NVFlare and can also integrate with Flower. Other frameworks such as OpenFL, FATE, or PySyft have not yet been tested with Apheris, but conceptually any server-based federated engine could be integrated.
Open-source federated learning frameworks like Kaapana or NVFlare have proven effective for multi-site hospital collaborations, but they primarily address where computation happens (keeping data on-site) without fine-grained control over what gets computed. Apheris adds a computational access governance layer: data custodians define per-asset policies that restrict which computations may run on their data, ensuring only approved, privacy-preserving workloads execute. The Gateway is designed to integrate into existing infrastructure with minimal deployment overhead, making it practical for cross-organizational collaborations where multiple independent parties need auditable control over how their data is used.

\section{Related Work}

Recent advances in multiple sclerosis (MS) research have increasingly leveraged federated learning \cite{mcmahan2017communication} to enable privacy-preserving analysis of multi-center clinical data. Personalized federated learning approaches have been proposed to improve predictive performance by adapting shared models to local distributions, using techniques such as selective parameter sharing and personalized fine-tuning \cite{Pirmani2025}. Federated learning has also been integrated into broader multi-layer data pipelines to facilitate large-scale collaboration and systematic data processing across institutions \cite{Pirmani2023}.  
Complementary approaches in MS data modeling employ Bayesian methods, machine learning techniques, and Common Data Model (CDM) based federated learning to harmonize heterogeneous real-world datasets and enhance predictive modeling \cite{Trojano2025}. In the context of imaging, federated learning has been applied to improve MS lesion segmentation across clinical sites, incorporating noise-resilient training and label correction to enhance segmentation performance \cite{BAI2024102872}.  
Moreover, explainable federated learning methods have been explored for MS detection and lesion localization, enabling interpretable models that provide insight into both prediction and spatial localization of disease features \cite{10.1007/978-3-032-11381-8_3}.

In this work, we use artificially simulated data to illustrate how federated image segmentation and federated analysis are already being applied in practice. In collaboration between Roche and Apheris, important contributions to MS-induced lesion segmentation \cite{10.3389/fneur.2025.1620469} and MS disease progression \cite{oh2023intonate} have already been realized within the INTONATE-MS consortium.

\section{Image analysis}
We now present a concrete showcase that demonstrates how these concepts can be applied in practice and begin with the image analysis. To this end, we fine-tuned an nnU-Net \cite{isensee2018nnunetselfadaptingframeworkunetbased,Isensee2021} in a federated learning setting using the two imaging datasets described below.
\subsection{Dataset Description}
The \texttt{mslesseg} \cite{Guarnera2025} dataset contains 115 NIFTI brain MRI scans from 75 patients, with three channels: T1, T2, and FLAIR. Each scan has 182 slices and an associated segmentation mask marking expert-annotated brain lesions for every slice. The 115 scans are split by patient ID into two subsets of 50 and 65 images.
Both datasets were further split into train and test such that we end up with 41 train and 9 test images on site-1 and 52 train and 13 test images on site-2.

\subsection{Federated Training}

To run the federated fine tuning on both image datasets we use Apheris Gateway and specify first a compute spec with the following dataset, model and research configuration
\begin{lstlisting}[language=Python]
compute_spec_id = compute.create_from_args(
    dataset_ids=dataset_ids,
    model_id="apheris-nnunet",
    model_version="0.28.0",
    client_memory=32000,
    client_n_cpu=14,
    client_n_gpu=1,
    server_memory=16000,
    server_n_cpu=7,
)
\end{lstlisting}

Once the compute spec is activated and running via Apheris CLI, jobs can be submitted to the compute spec that trigger the federated training. 
A typical training payload is shown below

\begin{lstlisting}[language=Python]
payload = {
    "mode": "training",
    "device": "cuda",
    "num_rounds": 30,
    "model_configuration": "2d",
    "dataset_id": 123
}

job.submit( payload,
    compute_spec_id=compute_spec_id, 
    verbose=True
)
\end{lstlisting}

\begin{figure}
    \centering
    \includegraphics[width=\linewidth]{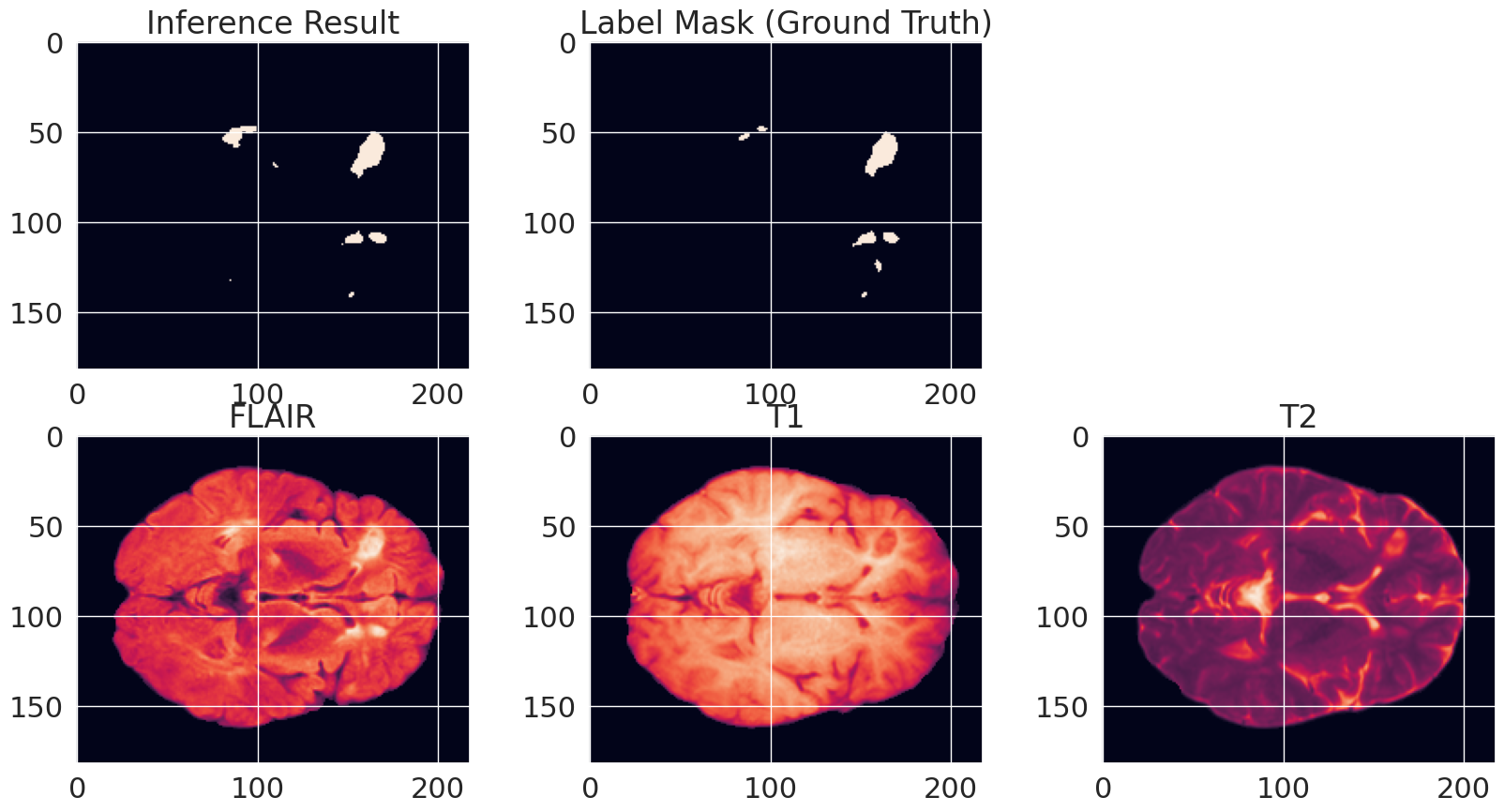}
    \caption{Single slice of image data point (bottom row) together with true and predicted segmentation mask (top row). The dice score on the selected slice is 0.85, for the overall image 0.67. In the bottom row MRI images on FLAIR, T1 and T2 channel are shown.}
    \label{fig:mri}
\end{figure}

Once the training job has finished, a model checkpoint can be downloaded via Apheris CLI and further be used for inference.
An inference job will produce for each image in the inference set an inferred segmentation mask per each slice of the MRI scan.
One slice of the raw data together with ground truth and inferred segmentation mask is shown in Figure \ref{fig:mri}.

The federated model training presented here is intended as a proof-of-concept to demonstrate the end-to-end workflow rather than to optimize model performance. Quantitative improvements achieved through federated fine-tuning within the INTONATE-MS consortium have been reported in~\cite{10.3389/fneur.2025.1620469}, where the federated nnU-Net model achieved dice scores ranging from 0.66 to 0.80 in the evaluations.
While the image data and ground truth are sensitive and not directly visible to the user, the inferred segmentation mask can either be returned to the user directly, if not considered as sensitive, or persisted on the gateway for further processing and aggregation, depending on the use case and model configuration.

\section{Basic statistics}
\begin{figure}[h!]
    \centering
    \includegraphics[width=\linewidth]{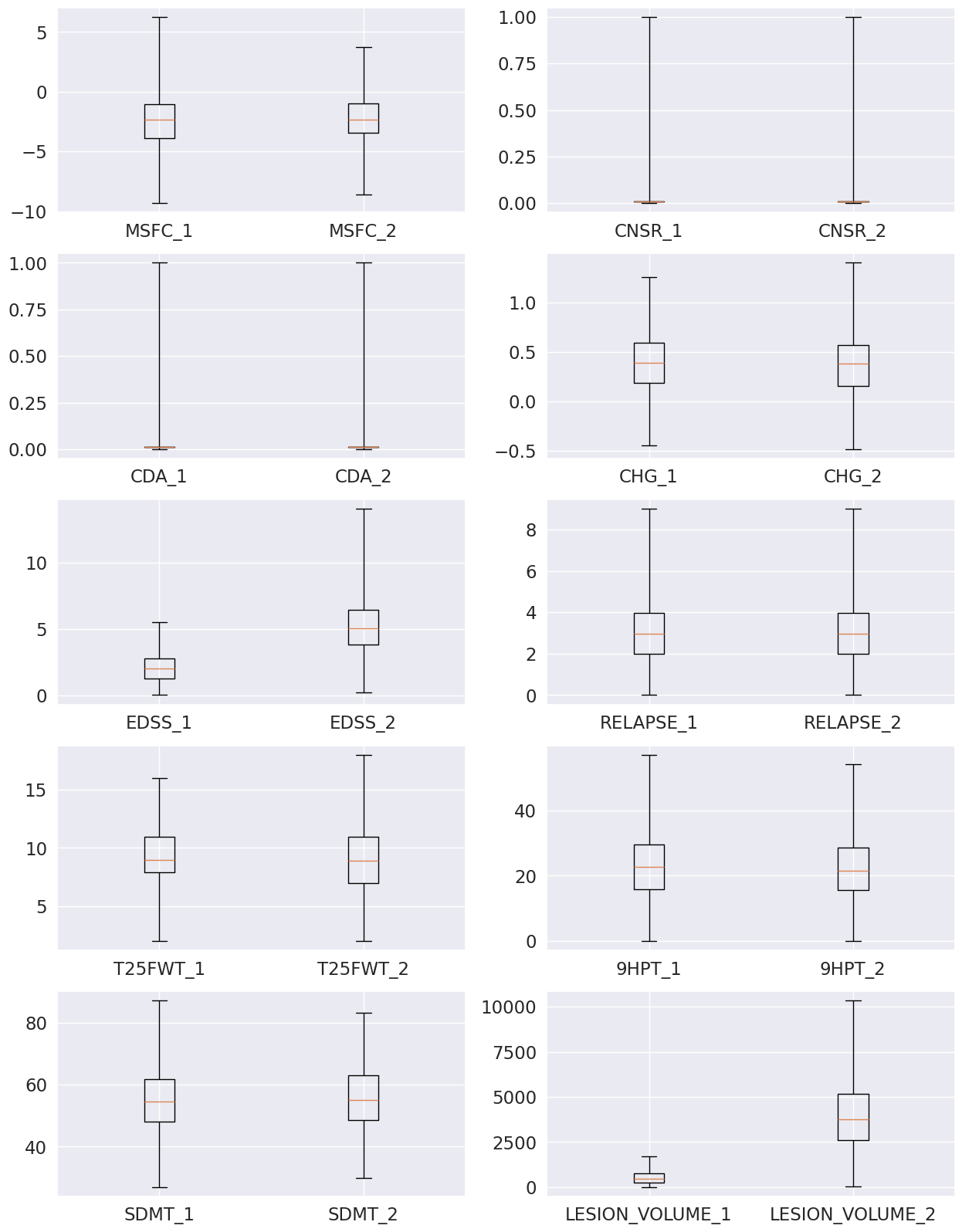}
    \caption{Boxplots of key variables for both sites with median, quartiles, and range shown.}
    \label{fig:eda}
\end{figure}

One piece of information derived from the inferred segmentation mask is the brain lesion volume. We assume this information is available in our clinical data tables and explore the relation to other clinical measurements.

We begin with a simple exploratory data analysis, starting with a \texttt{tableone} computation that provides descriptive statistics for a 
selection of numerical columns. The result is shown in Table \ref{tab:tableone}.

\begin{table*}[h!]
\centering

\begin{tabular}{l r r r r r r r r}
\toprule
Variable & n & Mean & SD & Min & Quartile 1 & Median & Quartile 3 & Max \\
\midrule
Symbol Digit Modalities Test (SDMT) 
 & 1386 & 55.56 & 10.15 & 27.00 & 48.60 & 54.60 & 63.00 & 87.00 \\

Annualized Change (CHG) 
 & 1386 & 0.39 & 0.31 & -0.48 & 0.18 & 0.38 & 0.59 & 1.40 \\

Expanded Disability Status Scale (EDSS) 
 & 1386 & 3.51 & 2.16 & 0.03 & 1.71 & 2.97 & 4.93 & 14.03 \\

Relapse Count (RELAPSE) 
 & 1386 & 3.11 & 1.82 & 0.00 & 1.98 & 2.97 & 3.96 & 9.00 \\

Censoring Indicator (CNSR) 
 & 1386 & 0.01 & 0.11 & 0.00 & 0.01 & 0.01 & 0.01 & 1.00 \\

Multiple Sclerosis Functional Composite (MSFC) 
 & 1386 & -2.33 & 2.02 & -9.33 & -3.72 & -2.32 & -1.07 & 6.26 \\

Timed 25-Foot Walk Test (T25FWT) 
 & 1386 & 8.91 & 2.53 & 2.00 & 6.96 & 8.88 & 10.96 & 18.00 \\

Nine-Hole Peg Test (9HPT) 
 & 1386 & 22.66 & 9.77 & 0.00 & 15.96 & 21.66 & 28.50 & 57.00 \\

Brain Lesion Volume 
 & 1386 & 2188 & 2136 & 1 & 414 & 1034 & 3722 & 10338 \\

Confirmed Disability Accumulation (CDA) 
 & 1386 & 0.0 & 0.2 & 0.0 & 0.0 & 0.0 & 0.0 & 1.0 \\
\bottomrule
\end{tabular}
\vspace{0.2cm}
\caption{The \texttt{tableone} function provides a summary of all key variables. The values were rounded to two decimal places depending on their magnitude, and in the case of Brain Lesion Volume, to whole numbers.}
\label{tab:tableone}
\end{table*}

We apply this analysis both in a federated manner to the entire collection of datasets and separately to each site in order to identify potential differences in data distributions or magnitudes.
The result is shown in Figure \ref{fig:eda} as boxplots. The comparison of both sites shows a small deviation in MSFC and T25FT distributions and a larger one for EDSS and Brain Lesion Volume. 

To get a clearer picture of how the different features in the data are connected, we take the correlation matrix into investigation.
Here, a separated analysis of both sites is compared to the federated analysis on the statistical ensemble.

\begin{figure*}[h!]
    \centering
    \includegraphics[width=0.9\linewidth]{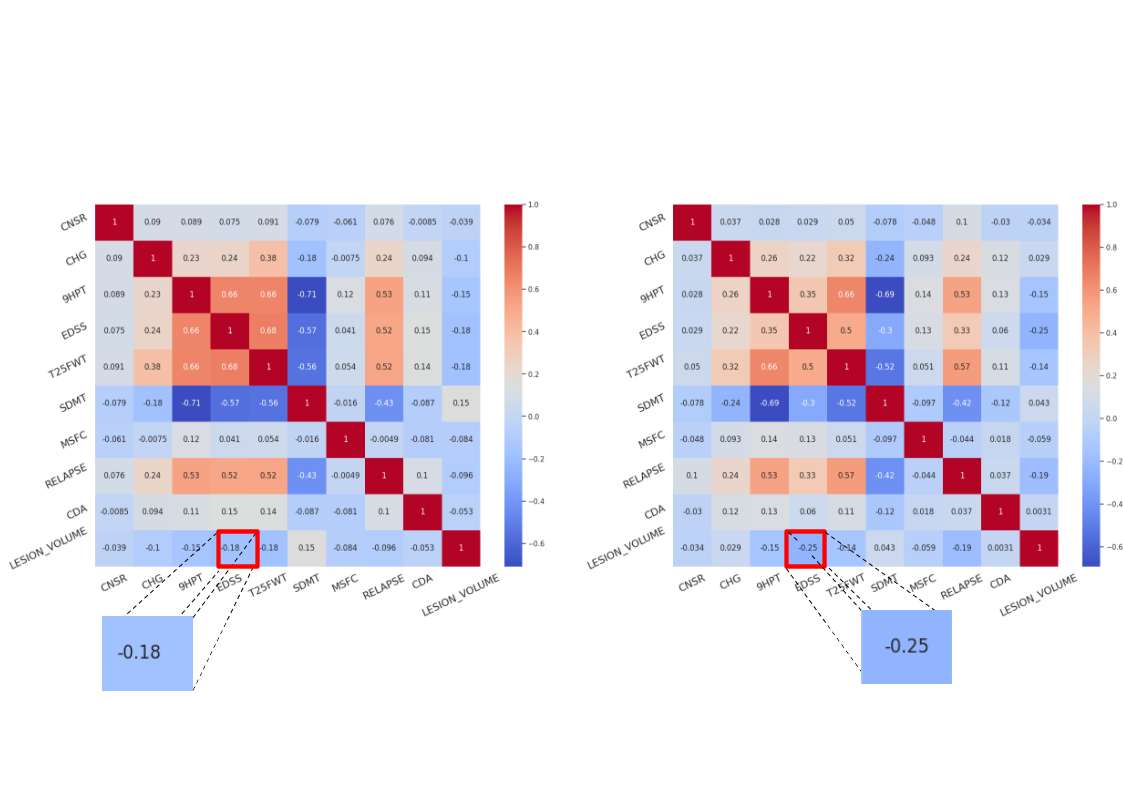}
    \caption{Correlation matrices for each site, computed independently, show negative correlations between brain lesion volume and EDSS.}
    \label{fig:corr-2-sites}
\end{figure*}

The correlation matrices computed at each site independently show a reasonably correlated block of the functional outcome measures of MS and apparently a negative correlation between EDSS and the volume of brain lesions (cf. Figure \ref{fig:corr-2-sites}).

\begin{figure}[h!]
    \centering
    \includegraphics[width=0.9\linewidth]{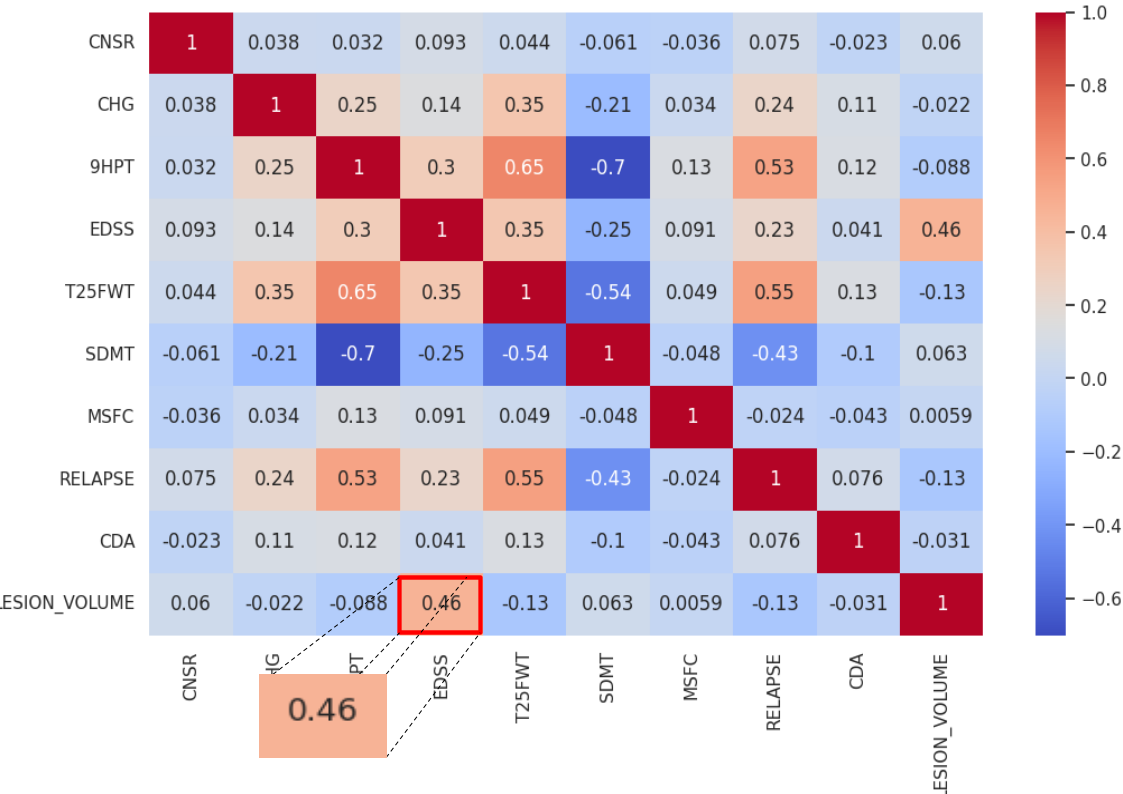}
    \caption{The federated correlation matrix based on the full data shows a positive correlation between brain lesion volume and EDSS.}
    \label{fig:fed-corr}
\end{figure}

The interpretation of this negative correlation is not immediately clear, so for context we consult the federated correlation matrix.
The federated analysis shows a markedly different relationship between the two variables compared to the individual dataset results. As seen in Figure \ref{fig:fed-corr} the two features exhibit a noticeable positive correlation.

This analysis across sites reveals that the negative correlations observed in the individual subsets may merely be an example of Simpson's Paradox\cite{Simpson1951Interpretation}, where correlations seen in different groups reverse when these groups are combined. This can be confirmed by examining the underlying data in an aggregated scatter plot. This is an intuitive way to compare the relationships observed in each site and the overall relationship seen across sites.

\begin{figure}[h!]
    \centering
    \includegraphics[width=0.9\linewidth]{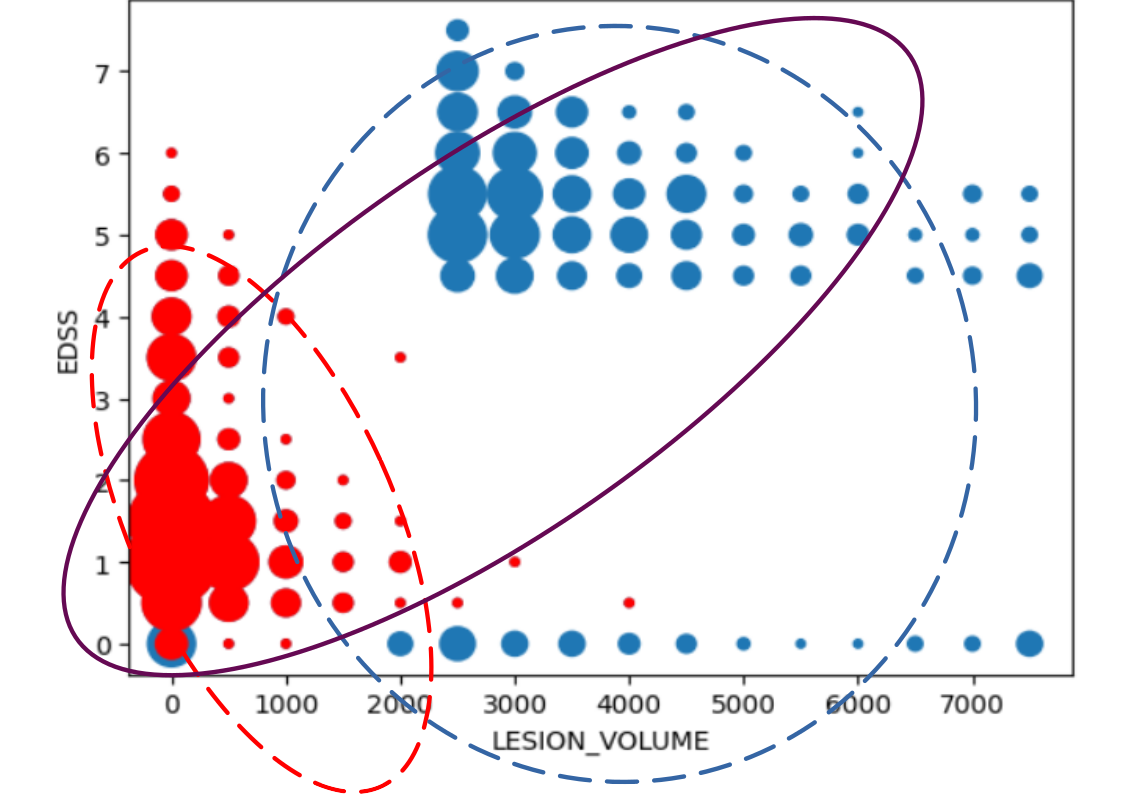}
    \caption{Scatter plot of discretized EDSS score vs. Brain Lesion Volume. Individual site results are shown in red/blue and exhibit a negative correlation, while aggregated results in purple show a positive correlation.}
    \label{fig:scatter}
\end{figure}

Because privacy constraints prevent us from inspecting individual data points, we define a box discretization over the dimensions Brain Lesion Volume and EDSS, and perform a federated computation to count how many data points fall into each box. These aggregated points can then be visualized with dot size depending on the number of observations in each box to enable us to clearly see in Figure \ref{fig:scatter} how the differing correlations arise. Note that care must be taken when choosing the binning (box size), as this may affect the visual perception of the correlation.

\section{Survival Analysis}

The previous section showcased how federated image models can help to detect brain lesions and how to assess their impact on other clinical measurements with the help of federated statistics.
To analyze the disease progression over time, this section focuses on survival analysis.

The first step is to prepare the date variables. Apheris preprocessing converts standard date formats into datetime objects that are easy to work with. A \texttt{tableone} summary of the date variables is shown in Table \ref{tab:eda-dates}.
\begin{table*}[h!]
\centering
\begin{tabular}{l r r r r r r}
\toprule
Variable  & n & Min & Quartile 1 & Median & Quartile 3 & Max\\
\midrule
Date of First Symptom (DTFSTSYM) 
& 1386 & 2022-01-02 & 2022-07-17 & 2023-01-01 & 2023-07-09 & 2023-12-31  \\

Diagnosis Date (DIAGDT) 
 & 1386& 2022-02-01 & 2022-08-17 & 2023-02-02 & 2023-08-04 & 2024-02-03 \\

Visit Date (VISITDT) 
 & 1386& 2022-02-26 & 2022-12-17 & 2023-06-08 & 2023-12-08 & 2024-12-09 \\

Treatment Start Date (TRTSDTC) 
 & 1386& 2022-03-02 & 2022-09-16 & 2023-02-25 & 2023-09-05 & 2024-03-07 \\
\bottomrule
\end{tabular}
\vspace{0.2cm}

\caption{\texttt{tableone} results on date columns}
\label{tab:eda-dates}
\end{table*}

\subsection{Kaplan-Meier Plots}
\begin{figure}
    \centering
    \includegraphics[width=0.8\linewidth]{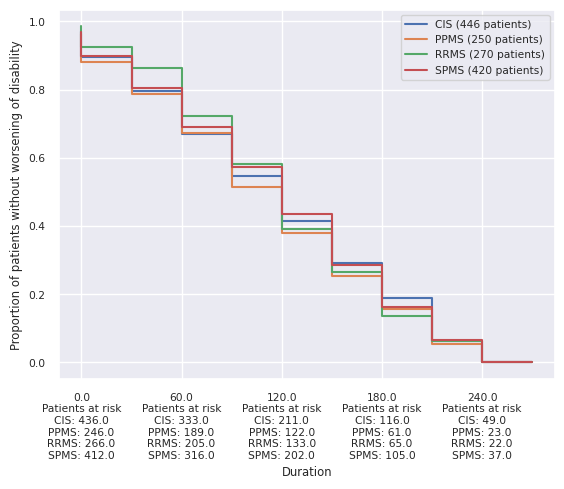}
    \caption{Kaplan-Meier plot for the event of EDSS exceeding 2 with time intervals used to preserve privacy.}
    \label{fig:km-surv}
\end{figure}

Datetime objects allow the calculation of date differences. The days since diagnosis are derived from the difference of Visit Date and Diagnosis Date and then used to compute a Kaplan-Meier survival function \cite{kaplan1958nonparametric}. In this instance, we are not analyzing patient mortality, but rather the probability of developing an MS-induced disability, as defined by worsening EDSS. We define the event $EDSS > 2.0$ and the Kaplan-Meier curve in Figure \ref{fig:km-surv} shows the "survival" probability of not worsening beyond this threshold. 
For this example, we defined the clinical event as $EDSS > 2.0$, representing the transition from subclinical neurological signs to measurable minimal disability. This threshold was selected to maximize the sensitivity of the survival analysis in detecting early disease progression and to provide a sufficient number of events for correlation with imaging biomarkers, which typically manifest early in the disease course.
In a real medical application, clinically established milestones such as $EDSS > 3.0$ or $4.0$ should be used instead.

\subsection{Cox Proportional Hazards Models}

We can now analyze individual patient data in a privacy-preserving way while simultaneously modeling disease progression at the population level. To unify these two perspectives, we conclude this case study with a Cox model \cite{cox1972regression}, which combines a baseline hazard function $H_0(t)$ with a covariate-dependent component $exp(\beta X)$, thus incorporating individual patient features as explanatory variables in the analysis. In the following we use the federated implementation of Cox regression as proposed by Andreux et al. \cite{andreux2020federatedcox}.

Together, these two components constitute the Cox proportional hazard function:
\begin{equation}
H ( t | X) = H_0(t) \cdot exp(\beta X).
\end{equation}

And from that we can derive the survival function as 

\begin{equation}
S(t|X) = exp\left(- H(t | X) \right)
\end{equation}

\begin{figure}
    \centering
    \includegraphics[width=0.8\linewidth]{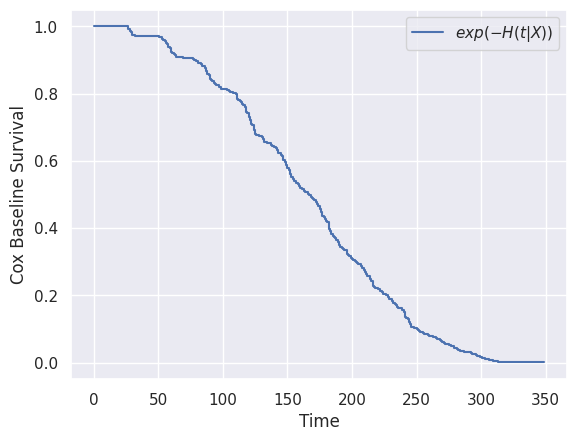}
    \caption{Survival function derived from Cox baseline hazard.}
    \label{fig:cox-baseline}
\end{figure}

The covariate coefficients in the Cox model quantify the influence of each feature on the hazard. In Table \ref{tab:cox-coef}, we present these coefficients and Figure \ref{fig:cox-baseline} shows the survival function derived from the baseline hazard. The coefficients normalized by the feature means can be interpreted as an indicator of feature importance.

\begin{table}[h]
\begin{tabular}{lr}
\toprule
{} &       $\beta$ \\
\midrule
Censoring Indicator (CNSR)                     & -0.0285 \\
Annualized Change (CHG)                        &  0.2133 \\
Nine-Hole Peg Test (9HPT)                      &  0.0004 \\
Timed 25-Foot Walk Test (T25FWT)               & -0.1683 \\
Symbol Digit Modalities Test (SDMT)            &  0.0154 \\
MS Functional Composite (MSFC) & -0.0307 \\
Relapse Count (RELAPSE)                        & -0.0478 \\
Confirmed Disability Accumulation (CDA)        &  0.5293 \\
Brain Lesion Volume                            & -0.000026 \\
\bottomrule
\end{tabular}
\vspace{0.2cm}

\caption{Coefficients of a Cox proportional hazards \\
model for the event of EDSS exceeding 2, \\ normalized 
by mean value.}
\label{tab:cox-coef}
\end{table}

\subsubsection{Cox Regression with Categorical Features}
Categorical variables can also be incorporated into the Cox proportional hazards model. To this end, an appropriate encoding is required. In our federated analysis, we first apply one-hot encoding using the Apheris preprocessing framework and then reduce the resulting high-dimensional feature space via principal component analysis (PCA) \cite{Jolliffe2002}.

\begin{figure}
    \centering
    \includegraphics[width=\linewidth]{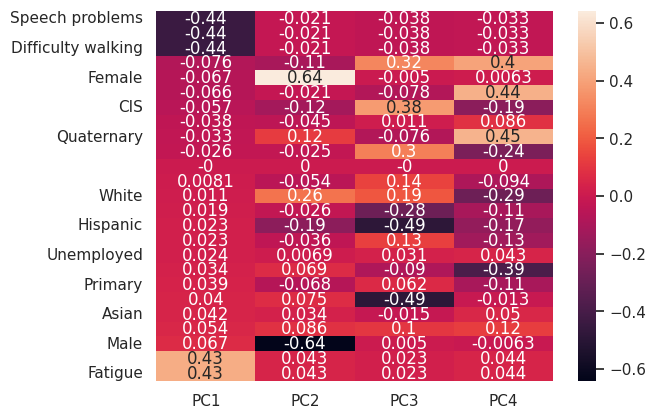}
    \caption{PCA transformation for one-hot encoded categoricals.}
    \label{fig:pca}
\end{figure}

As demonstrated earlier in this case study, Apheris Statistics enables federated computation of the covariance matrix, from which the transformation vectors for the PCA are derived. Figure \ref{fig:pca} illustrates the mapping from the one-hot encoded feature space to four principal components.

\section{Conclusion}

Federated learning offers a powerful paradigm shift for medical research. By enabling models to learn from distributed data sources without ever exposing sensitive patient information, it overcomes one of the largest barriers to collaborative healthcare innovation. As demonstrated, this approach allows researchers to uncover deeper insights from the full statistical ensemble rather than depending on fragmented, isolated data subsets. The result is stronger evidence, more robust models, and findings that better reflect real-world patient populations.

With purpose-built tooling for image segmentation, statistical modeling, and survival analysis, Apheris provides the technical foundation needed to drive this new generation of research. By combining privacy-preserving infrastructure with domain-specific analytics, Apheris empowers clinical researchers, data scientists, and pharma institutions to accelerate discovery.
The code for the showcase presented here is available as a notebook at https://www.github.com/apheris/simulated-brain-lesion-analysis.

\bibliographystyle{abbrv}
\bibliography{bibliography.bib}

@misc{Guarnera2025,
author = "Francesco Guarnera and Alessia Rondinella and Elena Crispino and Giulia Russo and Clara Di Lorenzo and Davide Maimone and Francesco Pappalardo and Sabastiano Battiato",
title = "{MSLesSeg: baseline and benchmarking of a new Multiple Sclerosis Lesion Segmentation dataset}",
year = "2025",
month = "6",
url = "https://springernature.figshare.com/articles/dataset/MSLesSeg_baseline_and_benchmarking_of_a_new_Multiple_Sclerosis_Lesion_Segmentation_dataset/27919209",
doi = "10.6084/m9.figshare.27919209.v1"
}

@ARTICLE{10.3389/fneur.2025.1620469,
  
AUTHOR={Hindawi, Sarah  and Szubstarski, Bartlomiej  and Boernert, Eric  and Tackenberg, Björn  and Wuerfel, Jens },
         
TITLE={Federated learning for lesion segmentation in multiple sclerosis: a real-world multi-center feasibility study},
        
JOURNAL={Frontiers in Neurology},
        
VOLUME={Volume 16 - 2025},

YEAR={2025},

URL={https://www.frontiersin.org/journals/neurology/articles/10.3389/fneur.2025.1620469},

DOI={10.3389/fneur.2025.1620469},

ISSN={1664-2295}}

@inproceedings{oh2025rwPIRMA,
  title        = {Integrating multicentre data to explore rwPIRMA: Results from the INTONATE-MS consortium},
  author       = {Oh, Jiwon and Smolders, Joost and Buijs, Frederik and Federer-Gsponer, Joel and zu Hörste, Gerd Meyer and Mamdani, Muhammad and Perrone, Christopher and Mowery,  Danielle L. and Kühnel,  Timm and van Tulder,  Koen and Testa,  Carolina and Elze,  Markus C. and Pedotti,  Rosetta and Kaczmarek,  Lukasz and Sharma, Vishakha and Tackenberg, Björn abd Bar-Or, Amit and Wiendl, Heinz},
  booktitle    = {ECTRIMS 2025},
  year         = {2025},
  address      = {Barcelona, Spain},
  note         = {Conference poster, INTONATE-MS consortium}
}

@misc{oh2023intonate,
  title        = {Utility and Implementation of a Federated Research Infrastructure to Assess Lack of Disease Stability as a Real-World Surrogate of {PIRA}, by Combining {MS} Clinical Trial and Real-World Cohort Data (The {INTONATE MS} Consortium)},
  author       = {J Oh and J Smolders and F Buijs and R Pedotti and F Dahlke and L Kaczmarek and A Kemmisetti and V Sharma and D Heinzmann and B Tackenberg and A Bar-Or and H Wiendl},
  year         = {2023},
  month        = {oct},
  organization = {Roche and Genentech},
  howpublished = {\url{https://medically.gene.com/global/en/unrestricted/neuroscience/ECTRIMS-2023/ectrims-2023-poster-oh-utility-and-implementation.html}},
  note         = {ECTRIMS 2023 Conference Poster}
}

@misc{Apheris_gateway,
  author       = {Apheris},
  title        = {Register \& unregister datasets},
  howpublished = {\url{https://www.apheris.com/docs/gateway/latest/data-custodian/register-and-unregister-datasets.html}},
  note         = {Accessed: 2025-12-04}
}

@misc{isensee2018nnunetselfadaptingframeworkunetbased,
      title={nnU-Net: Self-adapting Framework for U-Net-Based Medical Image Segmentation}, 
      author={Fabian Isensee and Jens Petersen and Andre Klein and David Zimmerer and Paul F. Jaeger and Simon Kohl and Jakob Wasserthal and Gregor Koehler and Tobias Norajitra and Sebastian Wirkert and Klaus H. Maier-Hein},
      year={2018},
      eprint={1809.10486},
      archivePrefix={arXiv},
      primaryClass={cs.CV},
      url={https://arxiv.org/abs/1809.10486}, 
}

@article{elkordy2023federated,
  title        = {Federated Analytics: A Survey},
  author       = {Elkordy, Ahmed Roushdy and Ezzeldin, Yahya H. and Han, Shanshan and Sharma, Shantanu and He, Chaoyang and Mehrotra, Sharad and Avestimehr, Salman},
  journal      = {APSIPA Transactions on Signal and Information Processing},
  volume       = {12},
  number       = {1},
  pages        = {e4},
  year         = {2023},
  doi          = {10.1561/116.00000063},
  publisher    = {Now Publishers, Inc.}
}

@inproceedings{mcmahan2017communication,
  title={Communication-Efficient Learning of Deep Networks from Decentralized Data},
  author={McMahan, Brendan and Moore, Eider and Ramage, Daniel and Hampson, Seth and y Arcas, Blaise Aguera},
  booktitle={Proceedings of the 20th International Conference on Artificial Intelligence and Statistics (AISTATS)},
  pages={1273--1282},
  year={2017},
  publisher={PMLR}
}

@article{kaplan1958nonparametric,
  title={Individual Nonparametric Estimation from Incomplete Observations},
  author={Kaplan, Edward L. and Meier, Paul},
  journal={Journal of the American Statistical Association},
  volume={53},
  number={282},
  pages={457--481},
  year={1958},
  doi={10.1080/01621459.1958.10501452},
  jstor={2281868}
}

@article{cox1972regression,
  title={Regression Models and Life-Tables},
  author={Cox, David R.},
  journal={Journal of the Royal Statistical Society: Series B (Methodological)},
  volume={34},
  number={2},
  pages={187--220},
  year={1972},
  publisher={Wiley}
}

@misc{andreux2020federatedcox,
  title={Federated Survival Analysis with Discrete‐Time Cox Models},
  author={Andreux, Mathieu and Manoel, Andre and Menuet, Romuald and Saillard, Charlie and Simpson, Chlo{\'e}},
  year={2020},
  note={International Workshop on Federated Learning for User Privacy and Data Confidentiality in conjunction with ICML 2020},
  howpublished={arXiv:2006.08997},
  url={https://arxiv.org/abs/2006.08997}
}

@article{Isensee2021,
  author  = {Isensee, Fabian and Jaeger, Paul F. and Kohl, Simon A. A. and Petersen, Jens and Maier-Hein, Klaus H.},
  title   = {nnU-Net: a self-configuring method for deep learning-based biomedical image segmentation},
  journal = {Nature Methods},
  volume  = {18},
  number  = {2},
  pages   = {203--211},
  year    = {2021},
  doi     = {10.1038/s41592-020-01008-z},
  url     = {https://doi.org/10.1038/s41592-020-01008-z},
  issn    = {1548-7105},
}

@book{Jolliffe2002,
  author    = {Jolliffe, I. T.},
  title     = {Principal Component Analysis},
  edition   = {2nd},
  series    = {Springer Series in Statistics},
  publisher = {Springer-Verlag},
  year      = {2002},
  isbn      = {978-0-387-95442-4},
  doi       = {10.1007/b98835},
}

@InProceedings{10.1007/978-3-032-11381-8_3,
author="Niro, Filomena
and Di Renzo, Miriam
and Agnello, Patrizia
and Petyx, Marta
and Ciaramella, Giovanni
and Martinelli, Fabio
and Cesarelli, Mario
and Santone, Antonella
and Mercaldo, Francesco",
editor="Rodol{\`a}, Emanuele
and Galasso, Fabio
and Masi, Iacopo",
title="A Privacy-Preserving Method for Explainable Multiple Sclerosis Detection Through Federated Machine Learning",
booktitle="Image Analysis and Processing - ICIAP 2025 Workshops",
year="2026",
publisher="Springer Nature Switzerland",
address="Cham",
pages="29--40",
isbn="978-3-032-11381-8"
}

@article{Pirmani2025,
  author  = {Pirmani, Ashkan and De Brouwer, Edward and Arany, Adám and Oldenhof, Martijn and Passemiers, Antoine and Faes, Axel and Kalincik, Tomas and Ozakbas, Serkan and Gouider, Riadh and Willekens, Barbara and Horakova, Dana and Havrdova, Eva Kubala and Patti, Francesco and Prat, Alexandre and Lugaresi, Alessandra and Tomassini, Valentina and Grammond, Pierre and Cartechini, Elisabetta and Roos, Izanne and Boz, Cavit and Alroughani, Raed and Amato, Maria Pia and Buzzard, Katherine and Lechner-Scott, Jeannette and Guimarães, Joana and Solaro, Claudio and Gerlach, Oliver and Soysal, Aysun and Kuhle, Jens and Sanchez-Menoyo, Jose Luis and Spitaleri, Daniele and Csepany, Tunde and Van Wijmeersch, Bart and Ampapa, Radek and Prevost, Julie and Khoury, Samia J. and Van Pesch, Vincent and John, Nevin and Maimone, Davide and Weinstock-Guttman, Bianca and Laureys, Guy and McCombe, Pamela and Blanco, Yolanda and Altintas, Ayse and Al-Asmi, Abdullah and Garber, Justin and Van der Walt, Anneke and Butzkueven, Helmut and de Gans, Koen and Rozsa, Csilla and Taylor, Bruce and Al-Harbi, Talal and Sas, Attila and Rajda, Cecilia and Gray, Orla and Decoo, Danny and Carroll, William M. and Kermode, Allan G. and Fabis-Pedrini, Marzena and Mason, Deborah and Perez-Sempere, Angel and Simu, Mihaela and Shuey, Neil and Singhal, Bhim and Cauchi, Marija and Hardy, Todd A. and Ramanathan, Sudarshini and Lalive, Patrice and Sirbu, Carmen-Adella and Hughes, Stella and Castillo Trivino, Tamara and Peeters, Liesbet M. and Moreau, Yves},
  title   = {Personalized federated learning for predicting disability progression in multiple sclerosis using real-world routine clinical data},
  journal = {npj Digital Medicine},
  volume  = {8},
  number  = {1},
  pages   = {478},
  year    = {2025},
  doi     = {10.1038/s41746-025-01788-8},
  url     = {https://doi.org/10.1038/s41746-025-01788-8},
  issn    = {2398-6352},
}

@article{Pirmani2023,
  author  = {Pirmani, Ashkan and De Brouwer, Edward and Geys, Lieven and Parciak, Tomasz and Moreau, Yves and Peeters, Liesbet},
  title   = {The Journey of Data Within a Global Data Sharing Initiative: A Federated 3-Layer Data Analysis Pipeline to Scale Up Multiple Sclerosis Research},
  journal = {JMIR Medical Informatics},
  volume  = {11},
  pages   = {e48030},
  year    = {2023},
  doi     = {10.2196/48030},
  url     = {https://medinform.jmir.org/2023/1/e48030},
}

@article{Trojano2025,
  author  = {Trojano, Marco and Iaffaldano, Pasquale and Copetti, Michele and Drahota, Jan and Forsberg, Lars and Mouresan, Elena F. and Pontieri, Luca and Spelman, Timothy and Toschi, Nicola and Butzkueven, Helmut and Glaser, Annika and Hillert, Jan and Horakova, Dana and Magyari, Melinda and Vukusic, Serge and Lucisano, Giancarlo and Kalincik, Tomas},
  title   = {Big multiple sclerosis data network: novel modelling approaches for real-world data analysis},
  journal = {Journal of Neurology},
  volume  = {272},
  number  = {12},
  pages   = {754},
  year    = {2025},
  doi     = {10.1007/s00415-025-13439-9},
  pmid    = {41206399},
}

@article{BAI2024102872,
title = {Improving multiple sclerosis lesion segmentation across clinical sites: A federated learning approach with noise-resilient training},
journal = {Artificial Intelligence in Medicine},
volume = {152},
pages = {102872},
year = {2024},
issn = {0933-3657},
doi = {https://doi.org/10.1016/j.artmed.2024.102872},
url = {https://www.sciencedirect.com/science/article/pii/S0933365724001143},
author = {Lei Bai and Dongang Wang and Hengrui Wang and Michael Barnett and Mariano Cabezas and Weidong Cai and Fernando Calamante and Kain Kyle and Dongnan Liu and Linda Ly and Aria Nguyen and Chun-Chien Shieh and Ryan Sullivan and Geng Zhan and Wanli Ouyang and Chenyu Wang}
}

@misc{Apheris2025FederatedDataNetworks,
  author       = {Hagestedt, Inken},
  title        = {What is a Federated Data Network and How Does it Support Cross‑Institutional Research?},
  howpublished = {Apheris blog},
  year         = {2025},
  month        = {April 24},
  note         = {Published December 3, 2024, last updated April 24, 2025},
  url          = {https://www.apheris.com/resources/blog/federated-data-networks},
}

@misc{AWS2025FederatedProteinLM,
  author       = {Choudhury, Olivia and Trautmann, Evelyn and Hales, Ian and Prieto, José-Tomás and Ratan, Ujjwal},
  title        = {Federated learning-based protein language models with Apheris on AWS},
  howpublished = {AWS for Industries blog},
  year         = {2025},
  month        = {August},
  day          = {11},
  url          = {https://aws.amazon.com/blogs/industries/federated-learning-based-protein-language-models-with-apheris-on-aws/},
  note         = {Accessed 2025},
}

@article{Simpson1951Interpretation,
  author  = {Simpson, Edward Hugh},
  title   = {The Interpretation of Interaction in Contingency Tables},
  journal = {Journal of the Royal Statistical Society, Series B},
  volume  = {13},
  number  = {2},
  pages   = {238--241},
  year    = {1951},
  doi     = {10.1111/j.2517-6161.1951.tb00088.x},
  jstor   = {2984065}
}

@article{oh2023utility,
  title={Utility and Implementation of a Federated Research Infrastructure to Assess Lack of Disease Stability as a Real-World Surrogate of PIRA, by Combining MS Clinical Trial and Real-World Cohort Data (The INTONATE-MS Consortium)},
  author={Oh, Jiwon and Smolders, Joost and Buijs, Frederik and Pedotti, Rosetta and Dahlke, Frank and Kaczmarek, Lukasz and Kemmisetti, Anil and Sharma, Vishakha and Heinzmann, Dominik and Tackenberg, B and others},
  journal={Multiple Sclerosis Journal},
  volume={29},
  year={2023}
}

@article{HAGESTEDT2024101077,
title = {Toward a tipping point in federated learning in healthcare and life sciences},
journal = {Patterns},
volume = {5},
number = {11},
pages = {101077},
year = {2024},
issn = {2666-3899},
doi = {https://doi.org/10.1016/j.patter.2024.101077},
url = {https://www.sciencedirect.com/science/article/pii/S2666389924002368},
author = {Inken Hagestedt and Ian Hales and Eric Boernert and Holger R. Roth and Michael A. Hoeh and Robin Röhm and Ellie Dobson and José Tomás Prieto}
}

\end{document}